\documentclass[prd,aps,epsf,floats,amsfonts,amssymb,amsmath,nofootinbib]{revtex4} 
\usepackage{graphicx,amssymb,amsmath}

\newcommand{\bsa}{\begin{subeqnarray}}
\newcommand{\esa}{\end{subeqnarray}}
\newcommand{\be}{\begin{equation}}\newcommand{\ee}{\end{equation}}
\newcommand{\bea}{\begin{eqnarray}}\newcommand{\eea}{\end{eqnarray}}
\newcommand{\brr}{\begin{array}}\newcommand{\err}{\end{array}}
\newcommand{\bit}{\begin{itemize}}\newcommand{\eit}{\end{itemize}}
\newcommand{\ben}{\begin{enumerate}}\newcommand{\een}{\end{enumerate}}

\def\lab{\label}

\def\lf{\left}

\def\non{\nonumber}

\def\rar{\rightarrow}
\def\ri{\right}\def\ti{\tilde}

\def\1{{_{1}}}\def\2{{_{2}}}

\begin{document}

\title{Linguistics and some aspects of its underlying dynamics}

\author{Massimo Piattelli-Palmarini${}^{\flat}$
and Giuseppe Vitiello${}^{\dag}$
}

\vspace{2mm}

\address
{${}^{\flat}$Department of Linguistics, University of Arizona, Tucson, AZ, USA\\
${}^{\dag}$Dipartimento di Fisica ``E.R.Caianiello'' and INFN,
Universit\`a di Salerno, Fisciano (SA) - 84084 Italy
}

\vspace{0.5cm}


\begin{abstract}
{\bf Abstract}:
In recent years, central components of a new approach to linguistics, the Minimalist Program (MP) have come closer to physics. Features of the Minimalist Program, such as the unconstrained nature of recursive Merge, the operation of the Labeling Algorithm that only operates at the interface of Narrow Syntax with the Conceptual-Intentional and the Sensory-Motor interfaces, the difference between pronounced and un-pronounced copies of elements in a sentence and the build-up of the Fibonacci sequence in the syntactic derivation of sentence structures, are directly accessible to representation in terms of algebraic formalism. Although in our scheme linguistic structures are classical ones, we find that an interesting and productive isomorphism can be established between the MP structure, algebraic structures  and many-body field theory opening new avenues of inquiry on the dynamics underlying some central aspects of linguistics.

\vspace{4.8mm}

{\bf Keywords}:  syntax, minimalist program, Labeling Algorithm, algebraic methods, coherent states, self-similarity, Fibonacci progression, deformed Hopf algebra

\end{abstract}

\maketitle

$$                  $$

\section*{Contents}

I. Introduction

II. The Minimalist Program

~~II.A. The X-bar structures and their self-similarity

~~II.B. The Interfaces

~~II.C. The manifold of concepts

~~II.D. Copies of lexical elements

III. Concluding remarks

Appendix A. A little bit of history

Appendix B. Some useful formulas on Fibonacci matrix

Appendix C. On the entropy and the free energy

References

\newpage

\section{Introduction}

The linguistic component of the present work is based on a vast body of theory and corpora of data spanning over 50 years, covering many languages and dialects. Initially called ``transformational grammar", as we will see, it is now called ``Generative Grammar". The founder of this linguistic domain and its main contributor has been, and still is, Noam Chomsky, but the community of researchers extends to some 2,000 members, in various parts of the world. In particular, our work deals with a relatively recent version of the theory (ever since 1995) called ``The Minimalist Program"~\cite{Chomsky1995},
and more particularly with a very recent further development (2012 to the present) that has brought, as we will see in a moment, linguistics even closer to physics. Numerous critiques of Generative Grammar, some of these claiming to be absolutely radical, have been expressed in the last 50 years or so, and several of these are quite recent. All of these have been persuasively rebutted already, frequently motivating the opponents to revise their critiques, trying different novel objections. This debate is embodied in a vast literature, out of which we indicate only some basic references\footnote{ A use-based explanation of language is offered
in~\cite{Tomasello2003} and~\cite{Bybee2007}, for counters see~\cite{Wexler2002} and~\cite{CRAIN,Pietroski}; 
for a statistical approach to syntax and language learning, see~\cite{Reali,Perfors,Christiansen}, for counters see~\cite{Berwick,Berwick2013}.}.
We will not go into this debate. We take our moves from the most recent developments of the Minimalist Program (MP), persuaded that we do not need excuses for doing this. We start this paper with a very short, essential historical reconstruction of the main quite recent developments of the theory, with special emphasis on the theory of syntax. The interested reader may find in the Appendix A a brief summary of developments of research in linguistics starting from the mid-Fifties. We then show how some features of the Minimalist Program are quite well suited to a mathematical representation in terms of algebraic methods and tools. This goes beyond a pure, although difficult, formal exercise, since it reveals the dynamics underlying aspects of the MP, which thus appears much richer than one might had suspected. Especially, it uncovers many contact points of the linguistic structure with concrete properties of nonlinear algebraic formalism  commonly used in the description of physical systems.
In other words, we find that an isomorphism can be established between the MP linguistic structure and many-body field theory. In our opinion, a very rewarding result, no matter from which standpoint one looks at, e.g. recognizing the deep dynamical processes underlying the MP linguistic structures, or, vice-versa, the linguistic content of the many-body formalism. The plan of the paper is the following. In Section II, the Minimalist Program is presented from the historical perspective. In the subsections II.A the X-bar trees, their self-similarity properties and their formalization is introduced. In the subsections II.B, II.C and II.D the interfaces, the manifold of concepts and the copies of lexical elements, respectively, are discussed. Section III is devoted to final remarks where comments on the entropy and the arrow of time are presented. Finally, the Appendix A with historical notations, Appendix B with properties of the Fibonacci matrix and Appendix C for the entropy and free energy definitions and derivations close the paper.

\section{The Minimalist Program}

From 1995 to 2012 the new development has been ``The Minimalist Program", an attempt to consider several components of the earlier version of the theory, called "modules" of Government and Binding (GB) Theory (see below and the Appendix A) not as explanations, but as explananda, under more abstract, simpler, more basic notions. Increasing emphasis is being put on ``third factors of language design''\footnote{The other two factors are: genetic predispositions and peculiarities of the local language that the child has to learn~\cite{Chomsky2005}.},
that is, principles that are not specific to language, nor specific to biological systems; basically, minimal (strictly local) search, minimal computation.  In other words, the physics and the mathematics of language. For a broader approach to language and language evolution see~\cite{Perlovsky1,Perlovsky2} and references therein.

The simplest operation now is binary Merge: take from the lexicon
$\alpha$  and $\beta$  and combine them into a binary un-ordered set $\{\alpha,\beta\}$ which is, as a whole, of the same category as one of them (the head).  For instance: $\{{}_{\alpha}\alpha,\beta\}$. {\it The smart man} is a man, {\it man}  is the head, and we have a Noun Phrase\footnote{An antecedent standard analysis has been that the head is the determiner ``the'' and the syntactic constituent is a Determiner Phrase (DP) dominating the NP. Noam Chomsky has expressed doubts about this (personal communication April 2014).};
{\it Had been bought} is an event of buying (the verb {\it buy} in the past tense is the head)\footnote{Again, an antecedent standard analysis posited that the head is the auxiliary ``had'' and its complement an aspect phrase.}, we have a Verb Phrase.
This set, as a whole, is then Merged with a third element from the lexicon $\gamma$  getting
$\{ {}_{\gamma}\gamma,\{{}_{\alpha}\alpha,\beta\}\}$. The new construct (Syntactic Object, SO) can have the category of $\gamma$. We do not need to go into the details here.

This binary Merge of larger and larger, more and more complex, syntactic objects, is, then, again recursively repeated until the whole sentence is terminated. The syntactic process, called ``derivation'' is similar to a proof ending when the sentence is terminated. In more complex sentences, with subordinates, relatives, embeddings, the process goes on until the derivation finally stops~\cite{Chomsky2001}. 

 There are intermediate cyclic points of derivational (computational) closure, called Phases\footnote{For greater clarity, we will use upper case P for Phases in syntax and lower case p for phases in physics.}. The syntactic derivation (the specific mental computation) stops when a Phase is reached, and then a higher Phase is opened. The process continues inside-out, building higher and higher components in the syntactic hierarchy. All these recursive operations are binary. Merge can create binary sets by pairing different lexical items or whole constituents  (e.g. $\{V, NP\}$ a verb and a noun phrase). This is called External Merge (EM). Internal Merge (IM) merges a set $X$ with a term $Y$ of $X$ (where a term is a subset of $X$ or a subset of a term of $X$), yielding $\{Y, X\}$ with two {\it copies} of $Y$, one the copy merged with $X$ and one the term $Y$ of $X$, which remains, because $X$ is unchanged by the operation. That's crucial.  It means that displacement yields copies, providing the basis for semantic interpretation (``reconstruction''). Syntactic movement (see the Appendix A) is now reduced to the operation of Internal Merge. Merge leaves the items being merged unaltered (this goes under the name of No Tampering Condition NTC)\footnote{The No-Tampering Condition (NTC) is crucial because the syntactic operation Merge cannot modify the lexical items it applies to. For instance, the sub-units that form a composite word like {\it undiagonalizable} ({\it un-, al,- ble}) are never treated separately by the derivation. These are, however, features contained in the lexicon relevant to the Labeling Algorithm. In this case, we have a ``bundle of features'' that signals ``this is an adjective''.}.

There is also ``pair-merge'', for asymmetric adjunction, when a sentence has some optional qualifications of the noun phrase (as in {\it the book of poems [with the glossy cover] [from Blackwell]}, where the expressions between square brackets are adjuncts)\footnote{Obviously this Merge operation must be asymmetric. We cannot have {\it the book with the glossy cover from Blackwell of poems}.}.   Finally, we have the Agree operation ({\it he goes, they go, he has gone, they have gone}) under conditions of minimal (strictly local) search. Syntax has no other component than these, there is nothing else in syntax, it is therefore called Narrow Syntax (NS).

 This is unlike the previous Theory of Government and Binding (GB)~\cite{Chomsky1981,Haegeman1991},  
where syntax had many components, many operators, many modules, many entities. These are now, in Minimalism, subsumed under the very basic operation Merge. The head gives the name to the constituent it generates (nouns to Noun Phrases, verbs to Verb Phrases and so on). More generally, we have an $\{H, XP\}$ construction, a Head and a Phrase. The $X$ in $X\,P$ is a generalization, meaning that it can be any phrasal category.

Some kinds of heads are intuitively transparent and have been standardly known in traditional linguistics (nouns, verbs, adjectives, prepositions), while others are of more abstract kind and their identification has been far from obvious or even counter-intuitive (complementizer, inflection, tense, negation and more). What were previously (in the Theory of Government and Binding) called ``empty categories'' (because they are not pronounced or written) are now simplified in terms of copies. Copies come for free, so to speak, because they are elements already present in previous steps of the derivation, for instance items extracted from the lexicon\footnote{Different languages treat the copies differently. In most languages only the higher copy is pronounced, but there are languages in which the lower copy is pronounced and also languages in which all copies are pronounced. In the latter case, this applies to "short" elements (equivalent to the English "who", "which" and similar), never to whole Noun Phrases.}.  The replacement of empty categories with copies of lexical elements is a clear advantage in simplification and has proved to be a legitimate move.

We need to specify that the general condition of ``strict locality'' applies to the structure of the sentence, not necessarily to what is, or is not, ``close'' on the surface of the sentence. We cannot bluntly just count the number of words separating the affected elements in the sentence, what counts are the number and kind of nodes separating the affected elements in the syntactic tree. This central property of syntax is called ``structure dependence'' and it constitutes a sharp departure from many old and new anti-generativist approaches to language based on statistics or conventions of use.

The Generative Grammar theory, perfected around the early Eighties as the Theory of Government and Binding (GB), enjoyed immense success, underwent many refinements. It allowed a deep analysis of many languages and dialects. The core of GB were several syntactic ``modules'' (sic), each taking care of a kind of syntactic operations (Case, Theta Criterion, Binding, Control etc.). (For a basic introduction, see~\cite{Carnie}.) 
It also turned out that all, absolutely all, Phrases had the same structure: the X-bar structure, which is recursive: an element of the structure (a node of the X-bar tree) can contain another X-bar structure, and so on recursively, indefinitely\footnote{The $X$ is a portmanteau symbol, covering all kinds of Phrases.}.   In Minimalism, the picture has been drastically reduced, gaining in elegance, simplicity and depth of explanatory power\cite{Krivochen}. We will go back to this in Appendix A.

Until 1995 all of this was the core of Generative Grammar. Perhaps, this is a good point where to insert in our presentation a first part of our algebraic formalization. In fact, we will see that we obtain in a straightforward way the recursivity, or self-similarity, of the X-bar structures.

\subsection{The X-bar structures and their self-similarity}

It has become standard for several years, in Generative Grammar, to construct syntactic trees that have only two branches departing from each node. This is, in fact, referred to as ``binary branching''~\cite{Kayne1984}. 
In fact, we have a collection of binary entities. Lexical items are represented, by useful convention, as $(+/-)$. For instance Nouns are $(+N, -V)$, Verbs as $(-N,+V)$.  This notation can be straightforwardly extended to Phrasal Heads $(+H, -C)$ and Complements $(-H,+C)$. In the syntactic derivation, we have Terminal nodes $(+T)$ and nonterminal nodes $(-T)$. Copies of lexical items, or of larger structures, in a sentence can be pronounced $(+Pr)$ or not-pronounced $(-Pr)$. Recursive applications of Merge may produce a Phase $(+Ph)$ or not $(-Ph)$.  As we just saw, the most basic syntactic operation, Merge, generates a binary set. This suggests to us to formalize the binary branching in terms of
standard mathematical formalism of vector spaces and matrix operations and multiplications.
Thus, let us start by considering the orthonormal basis of $2 \times 2$ matrices (or operators) given by the
three matrices $\sigma_1, \,\sigma_2$, $\sigma_3$ and the unit matrix $I$:
\bea
\sigma_1 =
\frac{1}{2}\left(
  \begin{array}{cc}
    0 & 1 \\
    1 & 0 \\
  \end{array}
\right)~, \quad  \sigma_2 =
\frac{1}{2}\left(
  \begin{array}{cc}
    0 & -i \\
    i & 0 \\
  \end{array}
\right)~, \quad \sigma_3 =
\frac{1}{2}\left(
  \begin{array}{cc}
    1 & 0 \\
    0 & -1 \\
  \end{array}
\right)~, \quad I =
\left(
  \begin{array}{cc}
    1 & 0 \\
    0 & 1 \\
  \end{array}
\right)~.
\eea
The $\sigma_1, \,\sigma_2$, $\sigma_3$ are usually called the Pauli matrices and play an important r\^ole in quantum theories. Here, however, we use them by exploiting solely their purely algebraic properties, in a classical, standard mathematical fashion. The vectorial space of states on which the matrices operate is built on the basis of the states $\left(
  \begin{array}{c}
    0 \\
    1 \\
  \end{array}
\right)$ and $\left(
  \begin{array}{c}
    1 \\
    0 \\
  \end{array}
\right)$, which we will denote by $|0\rangle$ and $|1\rangle$, respectively. The scalar product is denoted by $\langle i | j \rangle = \delta_{ij}$, $~i,j = 0,1$. Thus we consider the $SU(2)$ group and the associated $su(2)$ algebra given by the commutation relations, which in terms of the matrices
$\sigma^{\pm} = \sigma_1 \pm i \,\sigma_2$, are written as~\cite{Perelomov}
\begin{equation}
\label{su2} [\sigma_3, \sigma^\pm] = \pm \sigma^\pm \, , ~~~
[\sigma^-,\sigma^+] = - 2 \sigma_3~.
\end{equation}

We will assume to have a collection of $N$ two-level objects (``particles'' or ``lexical elements''), which in a  standard fashion are represented for each $i$ by the ``ground
states'' $|0 \rangle_i$ and ``excited states'' $|1 \rangle_i$, $i=1, 2, 3,...N $. We may also write
 $\sigma_{3i} = {1 \over 2} ( |1 \rangle_{ii}
\langle 1| - |0 \rangle_{ii} \langle 0|)$, with eigenvalues $\pm {1
\over 2}$, and
$\sigma_i^+=  |1 \rangle_{ii} \langle 0|$ and $\sigma_i^-=  |0
\rangle_{ii} \langle 1|$.
In the algebra Eq, (\ref{su2}), we then have  $\sigma^\pm = \sum_{i=1}^N \sigma_i^{\pm}$ and $\sigma_3 =
\sum_{i=1}^N \sigma_{3i}$.

Transitions between the two states $|0 \rangle_i$ and $|1 \rangle_i$ are generated by $\sigma_i^+= \left(
  \begin{array}{cc}
    0 & 1 \\
    0 & 0 \\
  \end{array}
\right)$~, and  $\sigma_i^-= \left(
  \begin{array}{cc}
    0 & 0 \\
    1 & 0 \\
  \end{array}
\right)$~:   $~\sigma_i^- |1 \rangle_i =|0 \rangle_i$,  $~\sigma_i^+ |0 \rangle_i = |1 \rangle_i$,  $~\sigma_i^- |0 \rangle_i = 0$ and $~\sigma_i^+ |1 \rangle_i = 0$.

In order to see how the ``binary Merge'' between two states  is generated, consider first  for simplicity  these two states $|0 \rangle$ and $|1 \rangle$. In full generality, they may represent two lexical elements or two levels of the same lexical element. In the following we will consider generalization to the collection of $N$ elements and restore the index $i$, which now for simplicity we omit. Start with $|0 \rangle$. Here and in the following we do not consider the (trivial) possibility to remain in the fundamental state $|0\rangle$ (which is dynamically equivalent to ``nothing happens''). The interesting possibility is the one of the excitation process from  $|0\rangle$ to $|1\rangle$. This is obtained by applying $\sigma^+$ to $|0\rangle$:
\be \lab{1}
 |0 \rangle ~~~\rar ~~~\sigma^+ |0 \rangle = |1 \rangle
\ee
which, by associating $0 \leftrightarrow |0 \rangle$ and $1 \leftrightarrow |1 \rangle$, may be represented as $0 ~~\rar ~~1  ~\equiv ~|1 \rangle$.
As a first single step  the state $|1 \rangle$ has been singled out.
After that we describe the ``action'' on that state by application of the sigmas.
Note that clarifying ``how to go'' from one step to the next one introduces the ``dynamics''.
We have thus an ``operatorial dynamics'' by which multiplicity of states are generated.

Consider now that $\sigma^+ \, \sigma^+ = 0 = \sigma^- \, \sigma^-$
and $\sigma^+ \, \sigma^- |1 \rangle = 1 |1 \rangle$ and  $\sigma^- \, \sigma^+ |0 \rangle = 1 |0 \rangle$. Therefore
the only possibilities to step forward of a single step is given in Eq. (\ref{1}),
%
%
and from there, one more single step is obtained as
\bea \lab{2a}
  &\nearrow& \sigma^- \sigma^+ |0 \rangle = |0 \rangle\\  
 |0 \rangle ~~~\rar ~~~\sigma^+ |0 \rangle = |1 \rangle ~~\rar ~~{} \non \\
&\searrow& \sigma^+ \sigma^- \sigma^+ |0 \rangle = |1 \rangle \lab{2b}
\eea
Note that application of $\sigma^+ \sigma^-$ is considered to produce a single step since it is equivalent to the application of the unit matrix $I$ to $|1 \rangle$. Note also that Eq.~(\ref{2a}) describes the ``decay process'' of the excited state $|1 \rangle$ to $|0 \rangle$. Eq.~(\ref{2b}) describes the ``persistence'' in the excited state, which represents a dynamically non-trivial possibility and thus we have to consider it. One more step forward leads us to

\bea \lab{12a}
  &\nearrow& \sigma^- \sigma^+ |0 \rangle = |0 \rangle ~~\rar ~\sigma^+ |0 \rangle ~~=~~~ |1 \rangle \\
|0 \rangle ~~\rar ~~\sigma^+ |0 \rangle = |1 \rangle ~~\rar ~~ \non \\
&\searrow& \sigma^+ \sigma^- \sigma^+ |0 \rangle = |1 \rangle_{\quad \large{\searrow \sigma^+ \sigma^- \sigma^+ |0 \rangle~= |1 \rangle}}^{\quad \nearrow \sigma^- \sigma^+ |0 \rangle ~~= ~~|0 \rangle} \lab{12b}\\
\non
\eea
\bea
~1  \qquad \qquad  \qquad  1 \qquad  \qquad  \qquad \qquad  \qquad \qquad    2    \qquad  \qquad \qquad  \qquad  ~~ 3 \label{8}
\eea
and so on. At each step, new branching points ${}^{\nearrow}_{\searrow} $ (new nodes of the X-bar tree) are obtained and the X-bar  tree is generated by the $SU(2)$ recursive dynamical process.

As said the application of $\sigma^+ \sigma^-$ in front of $\sigma^+ |0 \rangle$ is equivalent to multiply by $1$ the $\sigma^+$. In general, any power $(\sigma^+ \sigma^-)^n \, \sigma^+ |0 \rangle = 1 \times \sigma^+ |0 \rangle$, for any $n$.

The conclusion at this point is that we have the ``number of the states'' in these first steps in the sequence: $1 ~~1~~ 2 ~~3$, starting with $|0 \rangle$, [one state] , then in Eqs.~(\ref{1}) [one state],  (\ref{2a}) and  (\ref{2b}) [2 states], (\ref{12a}) and  (\ref{12b}) [3 states], respectively, (cf. Eq.~(\ref{8})).

From here, from the two $|1 \rangle$'s we will have in the next step two  $|0 \rangle$'s  and two $|1 \rangle$'s, and from the $|0 \rangle$ we will get one single $|1 \rangle$, in total 5 states:  $1 ~~1~~ 2~~ 3~~ 5$. We will get thus, in the subsequent steps, other states, and their numbers obtained at each step are in the Fibonacci progression ($\{F_n \}, F_{0} \equiv 0$)  with the ones obtained in previous steps. In general, suppose that at the step $F_{p+q}$ one has $p$ states $|0 \rangle$ and $q$ states $|1 \rangle$; in the next step we will have: $(p + q)\,|1 \rangle$ and $q\,|0 \rangle$, $F_{q + (p+q)}$. In the subsequent step: $(p + 2q)\,|1 \rangle$ and $(p + q)\,|0 \rangle$, a total of states $2p + 3q = (q + p + q) + (p + q)$, i.e. the sum of the states in the previous two steps, according to the rule of the Fibonacci  progression construction.

In conclusion, the X-bar tree (or F tree) has been obtained as a result of the $SU(2)$ dynamics, with the additional result that its recursivity or self-similarity properties turn out to be described by the Fibonacci progression.

It is interesting to remark that  $(\sigma^+ \sigma^-)^n \, \sigma^+ |0 \rangle = 1 \times \sigma^+ |0 \rangle$ can be thought of as a ``fluctuating'' process: $\sigma^+$ applied to the (ground) state $|0 \rangle$ excites it to $|1 \rangle$. Then $\sigma^-$ brings it down to $|0 \rangle$, and $\sigma^+$ again up to $|1 \rangle$: $\sigma^+ \sigma^-$ induces fluctuations $|1 \rangle~~\leftrightarrows ~~|0 \rangle ~~\leftrightarrows ~~|1 \rangle$ (through the ``virtual'' state $|0 \rangle$), this is the meaning of the fact above observed that $\sigma^+ \sigma^-$ is equivalent to 1 at any power $n$ when operating on $\sigma^+ |0 \rangle$. Similar fluctuations  can be  obtained by considering $(\sigma^- \sigma^+)^n \, \sigma^- |1 \rangle = 1 \times \sigma^- |1 \rangle$, i.e.  $|0 \rangle~~\leftrightarrows ~~|1 \rangle ~~\leftrightarrows ~~|0 \rangle$. This "fluctuating activity" corresponds, in the syntactic derivation, to successive applications of Merge. Simplifying a bit, when recursive Merge reaches the topmost node of a Phase, that is, a point of computational closure, everything underneath, in the tree, becomes off limit. The condition called ``Phase Impenetrability Condition'' (PIC)~\cite{Chomsky2001,Chomsky2000,Richards,Gallego}
specifies that nothing in a lower Phase is accessible to the syntactic operations that create the immediately higher Phase. The syntactic objects of the lower Phase and the lower Phase itself are dynamically ``demoted'' to a $|0 \rangle$ state. The ``fluctuating activity'' is also much suggestive when one thinks of the processes (of milliseconds or so) in the selection of lexical items and the recursive Merge of these into syntactic objects.

We also note that  a symmetric F three is obtained by exchanging at the start $|0 \rangle$ with $|1 \rangle$ and $\sigma^+$ with $\sigma^-$. Moreover we observe that a Jaynes-Cummings-like model Hamiltonian can be constructed generating the full set of $\sigma^+$ and  $\sigma^-$ products of any order, in all possible orderings compatible with the su(2) algebra (the products used above and leading, as we have seen, to the F progression). Thus our construction is a ``dynamical'' construction, a feature which certainly deserves much attention since we now have that the X-bar tree which plays  so a crucial role in the MP arises as a result of a {\it dynamical} model in linguistic, its recursive property being related to the self-similarity  property of the Fibonacci progression. The paramount importance of the Fibonacci progression in language has been stressed in~\cite{Medeiros,Idsardi,Piattelli2008}.

We close this subsection by observing that at any given step of the X-bar tree (the F tree), the simple knowledge of the state $|0 \rangle$ or $|1 \rangle$ is not sufficient in order to know its parent state in the previous step; we should also know which one is the branch we are on. This in part corresponds to the Phase Impenetrability Condition mentioned above and to one of the major problems in all of contemporary linguistic theory. In speaking and reading we proceed left to right, from the ``outside'' (the main sentence), to the ``inside'' (subordinate sentence), but the syntactic derivation proceeds from right to left, from inside out. This creates a conflict, that presumably the construction of Phases, of periodic points of closure, solves~\cite{Piattelli2008,Piattelli2004,Piattelli2005}.

While the tree construction (the ``way forwards'') is fully determined by the $\sigma$'s operations, the ``way backwards'' is not uniquely determined solely by the knowledge of the state $|0 \rangle$ or $|1 \rangle$. On the other hand, suppose one goes backwards of, say, $q$ steps starting from a given, say, $|1 \rangle$ (or $|0 \rangle$) . Then returning to such a specific state is no more guaranteed since at each branching point one has to chose which way to go (unless one keeps memory of its previous path, the Ariadne's thread...). In the syntactic derivation, ``forward'' consists in building further structure from the inside out, from right to left, proceeding upwards in the syntactic tree. The opposite, ``backwards'', consists in the derivation ``looking down'' to lower levels. The Phase Impenetrability Condition, as we have just seen, constrains this operation to a strict minimum. Omitting details, only the leftmost (and topmost) ``edge'' of the lower Phase is (quite briefly) still accessible to the operations building the next higher Phase.

The lesson is that, parameterizing by time the moving over the X-bar tree, time reversal symmetry is broken. We therefore need to deal with dissipative formalism. We will consider such a problem in the following. Before that we need to comment briefly in the following subsection on the ``interfaces'', namely the conceptual interpretive (semantic) system (CI) and the sensory-motor system (SM) to which Narrow Syntax has to make contact.

\subsection{The Interfaces}

Narrow Syntax has to make contact (has to interface) with two distinct systems: the conceptual interpretive (semantic) system (CI) and the sensory-motor system (articulation, auditory or visual perception) (SM). Language, for centuries, has been correctly conceived as sounds with meanings\footnote{Sound is the traditional expression, but we now know that it's unduly too restrictive: this should extend to gestures in sign languages (see the classic analysis of American Sign Language~\cite{Klima} 
and many studies ever since) to touch in deaf-and-blind
subjects~\cite{CChomsky1969, CChomsky1986}}, 
this is what Narrow Syntax and the interfaces give. But it's better now conceptualized as meanings with sounds, not sounds with meanings, because Narrow Syntax is optimized to interface with the conceptual interpretive system, not so much with the sensory motor system.
CI ``sees'' all copies, and interprets them, but at the SM interface only one copy is pronounced (usually the higher copy (see footnote n. 8)), while the other copy (or copies) remains silent (is deleted at SM).
\vspace{0,3cm}

- {\it Which books did you read [books]?}
\vspace{0,3cm}

The rightmost (lower, hierarchically) copy in English and in many other languages  is not pronounced (see the Appendix A). We see that ``copies'' now become important objects in the linguistic structure. We will show how this can be accounted in our modeling.

Until 2012, the ``optimality'' of Narrow Syntax with regard to the Conceptual Interpretive system was supposed to operate as follows: there are features that are ``meaningful'', that are called interpretable features, that CI can understand, and other features that are un-interpretable, meaningless. We comment more on this point in the Appendix A.
From 2012 to now, the bold hypothesis is that Merge does not form sets that have a category, not any more. It works freely and without constraints (a bit like Feynman's sum of all histories, before amplitudes give the wave function). It is ``only'' at the interface with CI (conceptual-interpretive) that categories are needed (CI needs labeled heads: which one is a verb, which one a noun, an adjective etc.)\footnote{It needs more than this: If $X\,P$ is a $V\,P$ at CI (the highest node, a Complementizer Phrase), then the mapping from narrow syntax to the Sensory-Motor system (externalization) must also know that it is a $V\,P$. Therefore labeling must be done at Transfer, so that the information goes to both interfaces.}. A minimal search process called {\it The Labeling Algorithm} is what does this job~\cite{Chomsky2013,Chomsky2014,Chomsky2015}.
In this framework, categorization and non-commutativity are only necessary at the CI interface. Order is important, obviously, at the Sensory-Motor interface (what to pronounce first, second etc. and what not to pronounce at all - deleted copies), but there is strong evidence that order doesn't appear at the CI interface. Order is probably a reflex of the Sensory Motor system, not feeding narrow syntax or CI\footnote{There are two notions of order to be taken into account: ordering of the syntactic operations and ordering of the items in the externalized linguistic expression. Here we have dealt with the latter, while a treatment of the first comes in what follows.}.   And categorization has to be the same at CI for interpretation at SM and for externalization.
 Today, some syntacticians try to shoehorn the previous analyses into this more stringent picture. Not everyone is persuaded that it can be done completely. But interesting explanations with elegant simplifications have been obtained
already~\cite{BerwickFriederici,Cecchetto,van Gelderen,Hornstein1999,Hornstein2005}
In essence: Explain and unify in terms of unconstrained Merge and the Labeling Algorithm many (ideally, all the) special properties of syntax. Nothing is invoked beyond the simplest computational operation Merge and reasonable interpretations of general principles of Minimal Computation (MC). It's Third Factors (physics) all the way.

\subsection{The manifold of concepts}

We are now ready to resume the discussion of the algebraic formalism. Our first task is to consider the whole set of $N$ elements (``particles'') introduced in the subsection II.A. We thus restore the subscript $i$ labeling our collection of $N$ elements. One may regard the collection of the associated states as the one at a given $F_n$ step, of high multiplicity $n$, in the Fibonacci tree. Furthermore, since $n$ can be as large as one wants, we may always have  ``since the beginning'' a direct product of a large number (in principle an infinite number, hence one needs field theories) of factor states, $\Pi_{\otimes, n,\infty}|0_i\rangle \equiv |0_1\rangle \otimes|0_2\rangle \otimes...|0_i\rangle \otimes... \equiv |0_1, 0_2,...0_i...\rangle $, $n$ of which (but not necessarily the first $n$ of them), in the $F_n$ step, will be  $|0\rangle$ (or $|1\rangle$).

Under the action of $\sigma^\pm$, the initial state assumed with all
particles in the ground state, $|0 \rangle_{\rm p}$, is driven
into the  states $|l \rangle_{\rm p}$ with
\begin{eqnarray} \label{L}
|l \rangle_{\rm p} &\equiv& \big[ \, |0_1 0_2 ... 0_{N-l}
\, 1_{N-l+1} 1_{N-l+2} ... 1_N \rangle + . . . \nonumber \\
&& \hspace*{1cm} + |1_1 1_2 ... 1_l \, 0_{l+1} 0_{l+2} ... 0_N \rangle
\, \big] / {\textstyle \sqrt{N
\choose l}} \, , ~~~
\end{eqnarray}
i.e. a superposition of all states with $l$ particles in $|1 \rangle$ and $N-l$ particles in $|0 \rangle$. The difference between excited and unexcited particles is given
by $\sigma_3$ since $_{\rm p} \langle l| \sigma_3 |l \rangle_{\rm
p} = l - {1 \over 2}N$. This quantity is called the order parameter. Its being non-zero signals that the rotational $SU(2)$ symmetry is broken\footnote{The phenomenon of spontaneous symmetry breakdown is throughly studied in many-body physics. For example, in the case of the electrical or magnetic dipoles, the order parameter provides the measure of the polarization or magnetization, respectively}.
For any $l$ we have
\begin{eqnarray} \label{rel}
&& \sigma^+ |l \rangle_{\rm p} = \sqrt{l+1} \, \sqrt{N-l}
\, |l+1 \rangle_{\rm p} \, , \nonumber \\
&& \sigma^-  |l \rangle_{\rm p} = \sqrt{N-(l-1)} \, \sqrt{l}
\, |l-1 \rangle_{\rm p} \, ,
\end{eqnarray}
showing that $\sigma^\pm$ and $\sigma_3$ are represented on $|l
\rangle_{\rm p}$ by the  so-called Holstein-Primakoff non-linear
realization \cite{Holstein,SUV,Vitiello,BJV}
\begin{eqnarray} \label{sigma}
&& \sigma^+ = \sqrt{N} S^+ A_{S} \, , ~~ \sigma^- = \sqrt{N} A_{S} S^-
\, , ~~~~ \nonumber \\
&& \sigma_3 = S^+ S^- - {\textstyle {1\over 2}} N
\end{eqnarray}
with $A_{S} = \sqrt{1 - S^+ S^-/N}$, $~~S^+  |l \rangle_{\rm p} =
\sqrt{l+1} \, |l+1 \rangle_{\rm p}~$ and $~~S^- |l \rangle_{\rm p} =
\sqrt{l}  \, |l-1 \rangle_{\rm p}$, for any $l$. The $\sigma$'s
still satisfy the su(2) algebra (\ref{su2}). However, in the limit of $N \gg
l$, Eqs.~(\ref{rel}) become
\begin{equation} \label{relc}
\sigma^\pm \, |l \rangle_{\rm p} =  \sqrt{N} \, S^\pm \, |l
\rangle_{\rm p}
\end{equation}
and thus $S^\pm = \sigma^\pm/\sqrt{N}$ for large $N$. For large $N$, the phenomenon of the contraction of the algebra  occurs~\cite{Wigner,Vitiello,Almut}. This means that in the present case in the large
$N$ limit the su(2) algebra (\ref{su2}) written now in terms of
$S^\pm$ and $S_3 \equiv \sigma_3$ contracts to the (projective)
e(2) algebra
\begin{equation} \label{e2}
[ S_3,S^\pm] = \pm S^\pm \, , ~~~ [S^-, S^+] = 1 \, .
\end{equation}

The result  Eq.~(\ref{e2}) is a central result. From a formal point of view it expresses the ``rearrangement'' of the su(2) algebra Eq.~(\ref{su2}) in the
e(2) algebra [Eq.~(\ref{e2})], which is isomorph to the Heisenberg-Weyl algebra, with $ S_3$ playing the role of the number operator and $S^\pm$ the role of ladder operators. The {\it rearrangement of symmetry} is a well known {\it dynamical} process~\cite{Vitiello,BJV,Umezawa:1993yq},  which occurs when there is spontaneous breakdown of symmetry characterized by a non-vanishing classical field called order parameter, with values in a continuous range of variability. In the present case, as mentioned above, the order parameter is given by $_{\rm p} \langle l| \sigma_3 |l \rangle_{\rm p} = l - {1 \over 2}N \neq 0$. One can show~\cite{SUV,Vitiello,BJV,Umezawa:1993yq} that when spontaneous symmetry occurs the space of the states of the system splits into (infinitely many) unitarily inequivalent representations of the algebra  Eq.~(\ref{e2}), that is it undergoes a process of {\it foliation}, splitting in many physically inequivalent subspaces, each one labeled by a specific value assumed by the order parameter. Each of these subspaces is a well defined vector space and represents a possible {\it phase} in which our system can live. Moreover, each of these {\it phases} is characterized by collective, coherent waves, represented by the ladder operators $S^\pm$. More specifically, the theorem can be proven~\cite{Goldstone}, which predicts the dynamical formation of long range correlation modes (the Nambu-Goldstone (NG) modes) when the symmetry gets spontaneously broken. In Generative Grammar the symmetry breaking phenomenon (the anti-symmetry of syntax, and the dynamic anti-symmetry of syntax) have been cogently argued for by Richard Kayne~\cite{Kayne}  and Andrea Moro~\cite{Moro}. 
This is in part why issues about the status of X-bar (as part of Narrow Syntax or as an emergent configuration of recursive binary Merge) have been recently debated~\cite{Chomsky2013,Chomsky2014} (see also Medeiros and Piattelli-Palmarini in preparation).
These NG modes, here represented by the ladder operators $S^\pm$, are the carrier of the {\it ordering} information through the system volume~\cite{SUV,Vitiello,BJV,Umezawa:1993yq}. {\it Order} thus appears as a collective dynamical property of the system which manifests in the limit $N \gg l$. The order parameter provides indeed a measure of the system ordering. Different degrees of ordering corresponding to different values, in a continuous range of variability, of the order parameter (different densities of the NG modes). Different values of the order parameter  thus denote different, i.e. inequivalent, phases of the system~\cite{SUV,Vitiello,BJV,Umezawa:1993yq}.

We thus realize that our system has undergone a formidable dynamical transition, moving from the regime of being a collection of elementary components (lexical elements) to  the regime of collective, coherent  $S^\pm$ fields. Our main assumption at this point is to identify a specific {\it conceptual, meaningful} linguistic content (a Logical Form LF\footnote{The notion of logical form (LF) as the "last" syntactic input to full meaning is now well consolidated in Generative Gramman (since~\cite{May}). A very simplified and intuitive example is given by what the following sentences have in common: Mary loves John. John is loved by Mary. It is Mary who loves John. It is John who is loved by Mary. In more complex sentences, the derivation of LF proceeds by steps, from the inside out, Phase by Phase.}) with the  collective coherent phase associated to a specific value of the order parameter. The semantic level, characterized by a {\it continuum} of concepts or meanings (the ``manifold of concepts''),  thus emerges as a dynamical process out of the syntactic background of lexical elements, in a way much similar (mathematically isomorph) to the one by which macroscopic system properties emerge as a coherent physical phase out of a collection of elementary components at a microscopic (atomistic) level in many-body physics~\cite{BJV,Umezawa:1993yq}.

In conclusion, we can now give a quantitative characterization of the ``interfaces'' where the Narrow Syntax has to make contact with the Conceptual Interpretative (CI) system: interfaces are met when the spontaneous breakdown of symmetry is met, i.e. as the limit $N \gg l$ is approached. It is there that a specific meaning or ``concept'' arises from a ``continuous'' context of possible concepts by selecting out one representation of the algebra from many of them ``unitarily inequivalent'' among themselves (each corresponding to a different concept)~\cite{Vitiello:1995}. The concept  appears at that point as a collective mode, not a result of associative process pulling together bits and little lexical pieces, words etc.. The collectiveness comes  from the ``phase coherence'', whose carriers are the collective NG $S^\pm$ fields. We also understand why ``only at the interfaces the  issues of ordering become relevant'' (cf. previous subsection). Order indeed is lack of symmetry and it can only appear when this is spontaneously broken.

For the same reason, categorization and non-commutativity (and order) are only necessary at the CI interface. Indeed, only at the large $N$ limit CI needs labeled heads: which one is a verb, which one a noun, an adjective etc..
We have seen that the formal construction of the binary Merge does not require labeled structures (Noun, Verb, Adjective, Preposition etc.). The necessity of labeling ({\it The Labeling Algorithm}) only arises at the interface with meaning. Interpreting a Determiner Phrase ({\it the dog, many boys, most books}) or a Verb Phrase ({\it was going to Rome, had bought the book}) or a Complementizer Phrase ({\it that I know him, whether it's wise, who did that�}) is a necessity for the Conceptual Intentional system, with the formal label of a syntactic object triggering different Intentional landscapes.
Once the Narrow Syntax has made contact with CI system, through the action-perception cycle~\cite{Vitiello:1995} of the cortex dynamics, the sensory-motor system (SM) gets also involved and therefore the linguistic structures can be externalized, allowing to communicate to other speakers all the required subtleties of meaning.

The formalism here presented thus endorses Chomsky's thesis that Merge is unconstrained, and that issues of labeling (headedness, categorization of lexical items) and ordering only arise at the interfaces of Narrow Syntax with the Conceptual-Intentional (CI) system and the Sensory-Motor (SM) system.

\subsection{Copies of lexical elements}

We now consider the feature of  the copies of lexical elements in the Minimalist Program (MP).

At the end of subsection II.A, we have observed that time-reversal symmetry is broken moving along the X-bar tree. We express this by saying that the X-bar structures are dissipative structures. From the standpoint of the algebraic formalism, this means that we have to set up a proper mathematical scheme, which is achieved by doubling the system degrees of freedom~\cite{Celeghini1992}. This goes as follows.

The doubling of the degrees of freedom of the system, namely of the algebra mapping ${\it \cal A} \rar {\it \cal A} \times {\it \cal A}$, is a natural requirement to be satisfied when one has to consider, for example, the total energy of a system of two identical particles, ${\it \cal E}_{tot} = {\it \cal E}_1 + {\it \cal E}_2$, or their total angular momentum $L_{tot}  = L_1 + L_2$.  These sums denote
the Hopf coproducts ${\it \cal E}_{tot} = {\it \cal E} \times 1 + 1 \times {\it \cal E}$ and $L_{tot} = L \times 1 + 1 \times L$, respectively, which are commutative under the exchange of the two considered particles. Most interesting is the case of two elements (or particles) which cannot be treated on the same footing, as it happens when dealing, for example, with open or dissipative systems (e.g. finite temperature systems), where the system elements cannot be exchanged with the elements of the bath or environment in which the system is embedded, or as in the present case of linguistics where, at the syntactic and semantic levels,  lexical elements and conceptual contents cannot be simply interchanged. In these cases we need to consider $q$-{\it deformed} Hopf algebras with {\it noncommutative} Hopf coproducts  $\Delta A_{q} = A \times q + q^{-1} \times A \equiv A \,q \,+ \, q^{-1}\, {\tilde A}$~\cite{PLATh}, with the operator (matrix) $A \in {\cal A}$ and $q$ a number chosen on the basis of some mathematical constraint on which we do not need to comment here. The ``tilde'' operators ${\tilde A}$ thus denote the doubled operators in the doubling of the algebra ${\it \cal A} \rar {\it \cal A} \times {\it \cal A}$.  We stress that the introduction of the tilde-modes is in no way by hand or ad hoc. Their presence is implied by the very same Hopf algebra structure~\cite{BJV}.
Note that for simplicity we are omitting subscripts denoting the momentum $\bold k$  of the $A$ (and ${\tilde A}$) modes, e.g.  $A_{\bold k}$, as far as no misunderstanding occurs.

Summarizing,  we have the ``copies''  ${\tilde A}$ of the operators  $A$, the Hopf doubling of the algebra $A \rightarrow \,\{A,  \,  {\tilde A}\}$ and of the state space ${\cal F} \rightarrow {\cal F} \times {\tilde{\cal F}}$. The operators $A$ and  ${\tilde A}$ act on ${\cal F}$ and ${\tilde{\cal F}}$, respectively, and commute among themselves.

For notational simplicity from now on we will denote by $A$ and $A^{\dag}$ the operators $S^-$ and $S^+$ in Eq.~(\ref{e2}), respectively (as usual, the symbol $\dag$ in $A^{\dag}$ denotes the hermitian conjugate matrix, namely the transpose and complex conjugate of the matrix representation of A). Thus, the doubling process implies that correspondingly we also have ${\tilde S}^-$ and  ${\tilde S}^+$, which will be denoted as  ${\tilde A}$ and ${\tilde A}^{\dag}$, respectively.

In the following we work with $q(\theta) = e^{\pm\,\theta}$, which means that hyperbolic sine, cosine and tangent will be involved in the computations. By proper algebraic operations (see \cite{BJV,PLATh} for the technical details)  one then obtains   the  operators $A(\theta), \, {\tilde A}(\theta)$ and the so-called Bogoliubov transformations:
\bea \label{bog1}
A(\theta) &=& A \cosh \, \theta   - {\tilde A}^{\dag} \sinh \, \theta , \\ \lab{bog2}
{\tilde A}(\theta) &=& {\tilde A} \cosh \, \theta   -  A^{\dag} \sinh \, \theta .
\eea
The canonical commutation relations (CCR) are
\bea \label{ccr}
[A(\theta), A(\theta)^{\dag}] = 1 , \qquad  [{\tilde A}(\theta), {\tilde A}(\theta)^{\dag}] = 1 ,
\eea
All other commutators equal to zero. The Bogoliubov transformations Eqs.~(\ref{bog1})  and (\ref{bog2}) provide an explicit realization of the doubling or ``copy'' process discussed above.

Denote now by $|0\rangle \equiv |0\rangle \times |0\rangle$  the state annihilated by $A$ and $\tilde A$: $A |0\rangle = 0 = {\tilde A} |0\rangle$ (the vacuum state).
Note that  ${A}(\theta)$ and ${\tilde A}(\theta)$ do not   annihilate $|0\rangle$. The state annihilated by these operators is~\cite{BJV,Umezawa:1993yq,Celeghini1992}
\be \lab{vtheta}
|0 (\theta)\rangle_{\cal N} = e^{i \sum_{\bold k} \theta_{\kappa} G_{\bold k}}|0\rangle =
\prod_{\bold k} \frac{1}{\cosh \, \theta_{\kappa}} \exp(\tanh \theta_{\kappa} A^{\dag}_{\kappa}{\tilde A}^{\dag}_{\kappa} )\, |0\rangle ,
\ee
where the subscript $\bf k$ has been restored. The meaning of the subscript $\cal N$ is clarified below, $\theta$ denotes the set $\{\theta_{\kappa}, \forall {\bf k}  \}$ and  $|0(\theta) \rangle_{\cal N}$ is a well normalized state:  ${}_{\cal N}\langle 0(\theta)|0(\theta) \rangle_{\cal N} = 1$\footnote{The vacuum $|0 (\theta)\rangle_{\cal N} $ turns out to be a generalized $SU(1,1)$ coherent state of condensed couples of  $A$ and $\tilde A$ modes~\cite{Celeghini1992,Perelomov}, which  are entangled modes in the infinite volume limit.}, \footnote{In language, in first approximation, the vacuum state is silence. Just like in the present algebraic formalism, there are many kinds of silence. Not only how a silence gap is interpreted in the unfolding of a conversation, but in a more specific and more technical sense. There is, literally, a syntax of silence~\cite{Merchant} 
in linguistic constructions called ellipsis ({\it Mary bough a book and Bill did} ${}_{----}$ {\it too}) and sluicing ({\it Ann danced with someone but I do not know who} ${}_{----}$). Jason Merchant and other syntacticians and semanticists have persuasively shown that what can be omitted is never just an arbitrary ``bunch of words'', but an entire syntactic constituent (an entire Verb Phrase, most frequently). The syntax and semantics of several well defined but unpronounced elements has been part of the theory since the beginning of Generative
Grammar~\cite{Chomsky1955}.}.  
The operator $\exp \,( i \sum_{\bold k} \theta_{\kappa} G_{\bold k})$, with   $G_{\bold k} \equiv -i\,(A^{\dag}_{\bold k}{\tilde A}^{\dag}_{\bold k} \, - \,A_{\bold k}{\tilde A}_{\bold k} )$, is the generator of the Bogoliubov transformations (\ref{bog1}) and (\ref{bog2}) and of the state $|0 (\theta)\rangle_{\cal N}$ as shown in Eq.~(\ref{vtheta}).

One can show~\cite{Celeghini1992,PLATh,Blasone:1998xt} that   ${}_{\cal N} \langle 0|0(\theta) \rangle_{\cal N}  \rightarrow 0$ and ${}_{\cal N} \langle 0(\theta')|0(\theta) \rangle_{\cal N}  \rightarrow 0$, $\forall \theta \neq \theta'$,  in the infinite volume limit $V \rightarrow \infty$. Thus we conclude that the state space splits in infinitely many physically inequivalent representations in such a limit, each representation  labeled by a $\theta$-set $\{ \theta_{\bold k} = \ln \, q_{\bold k}, \, \forall {\bf k} \}$.  This is the $q(\theta)$-{\it foliation} process of the state space. For what said in subsection II.A and II.C, in the present case of linguistics this represents the process of generation of the manifold of concepts. It is a dynamical process since  the generator $G_{\bold k}$ is essential part of the system Hamiltonian~\cite{Celeghini1992}.
In our present case, ``physically inequivalent'' representations means that the ``manifold of concepts'' is made of ``distinct'', different spaces, each one representing a different ``concept'' (in language we have the Logical Forms (LFs) composing the global Logical Form of the entire sentence), here described as the coherent collective mode generated through the X-bar tree as illustrated in subsection II.C. These spaces (concepts) are protected against reciprocal  interferences since the spaces are ``unitarily inequivalent'', i.e. there is no unitary operator able to transform one space in another space~\cite{Vitiello:1995,Vitiello:2001}, which corresponds to the fact that syntactic Phases cannot be commingled, nor "reduced" one into the other. Phases are, as we said above, mutually impenetrable. In practice, however, the unitary inequivalence is smooted out by realistic limitations, such as, for example, the impossibility to reach in a strict mathematical sense the $V \rightarrow \infty$ limit (i.e. the ``infinite number''  of lexical elements or the theoretically infinite number of choices for the co-referentiality indices in the logical form of even the simplest sentences)\footnote{One of the leaders in the semantics of natural languages (Professor Irene Heim of MIT), wrote:
``{\it We just focused on a particular logical form that grammar provides for the sentence "She hit it" [...] But there are infinitely many others, since the choice of indices is supposed to be free. So} [the simple logical form there reported] {\it represents really only one of many readings that the sentence may be uttered with.}''(~\cite{Heim}  pag 232).}. 
Thus, realistically, we may also move from concept to concept in a chain or trajectory going through the manifold of concepts~\cite{Vitiello:1995,Vitiello:2004,VitielloDiss:2004,FreemanVitiello:2006,FreemanVitiello:2008,CapolupoPLR}. This corresponds to the compositionality of meanings, when the syntactic derivation proceeds "upwards" (that is: forward) from the lower Phases to the higher Phases, from local Logical Forms to the composition of more inclusive Logical Forms. Remarkably, these trajectories in the unitarily inequivalent spaces (the ``manifold of concepts'') have been shown to be classical chaotic trajectories~\cite{Vitiello:2004}.

In order to better understand the role played by the ``tilde copies'', $\tilde A$, it is interesting to compute
$N_{A_{\bf k}} = A^{\dag}_{\bold k} A_{\bold k}$ in the state $|0(\theta) \rangle_{\cal N} $
\be \lab{numop}
{\cal N}_{A_{\bf k}}(\theta) \equiv {}_{\cal N} \langle 0(\theta)| A^{\dag}_{\bold k} A_{\bold k}|0(\theta) \rangle_{\cal N}  =
{}_{\cal N} \langle 0(\theta)|{\tilde A}_{\bold k}(\theta) {\tilde A}^{\dag}_{\bold k}(\theta)|0(\theta) \rangle_{\cal N}   = \sinh^{2} \, \theta_{\bold k}.
\ee
From this we see that for any $\bf k$ the only non-vanishing contribution to the number of non-tilde modes ${\cal N}_{A_{\bf k}}(\theta)$  comes from the tilde operators, which can be expressed by saying that these last ones constitute the dynamic {\it address} for the non-tilde modes (the reverse is also true,  the only non-zero contribution to ${\cal N}_{{\tilde A}_{\bf k}}(\theta)$ comes from the non-tilde operators).

In conclusion, the physical content of
$|0(\theta) \rangle_{\cal N} $ is specified by the $\cal N$-set $\equiv \{ {\cal N}_{A_{\bf k}}(\theta), {\cal N}_{A_{\bf k}}(\theta) = {\cal N}_{{\tilde A}_{\bf k}}(\theta), \forall {\bf k}  \}$, which is called the {\it order parameter}. It is a characterizing parameter for the vacuum
$|0(\theta) \rangle_{\cal N} $ and explains the meaning of the $\cal N$ subscript introduced above.

All of this sheds some light on the relevance of ``copies'' in the MP. In some sense they are crucial in determining (indeed providing the  address of) the whole conceptual content of the considered linguistic structure. They provide the {\it dynamic reference} for the non-tilde modes. Un-pronounced copies, being silent, do not reach the Sensory Motor system, but they are crucially interpreted by the Conceptual Intentional system. They are necessary to the understanding of the meaning of what is actually pronounced. Remarkably, they are ``built in'' in the scheme here proposed; they are not imposed by hand by use of some constraint ``external'' to the linguistic system. It is in this specific sense that we speak of ``self-consistency'': our formal scheme is computationally (logically) self-contained. Perhaps the real power of the linguistic tool available to humans consists in such a specific feature.

\section{Concluding remarks}

The essence of the contribution we propose in this paper for the understanding and the  physical modeling of the Minimalist Program consists in having pointed out the {\it dynamical} nature of the transition from a numeration of lexical items to syntax and from syntax to the logical form (LF) of the sentence and from LF to meaning.
This has brought us to the identification of the manifold of concepts, to the self-similar properties of the X-bar trees and of their dissipative character (breakdown of the time-reversal symmetry), to the role of the copies in the conceptual interpretative system CI. The Hopf algebra structure has shown that the doubled tilde operators, which we have seen to play the role of the copies in the CI system, are ``buit in'' in the computationally self-contained algebraic scheme. These copies or tilde modes have been recognized to provide the  {\it dynamical reference} (the ``address'') of the non-tilde modes. The result is the logical self-consistency (inclusion of the {\it reference} terms) of languages.

We have also pointed out the mechanism of the foliation of the space of the states, out of which  the great richness of the  conceptual content, the ``multiplicity'' of inequivalent meanings (LF) emerges (see the comments following Eq.~(\ref{numop}) and the remark by Heim in the footnote 17). In this connection, we would like to call the attention of the reader on a further aspect of the scheme we propose in order to model some features of the MP, namely on its intrinsic thermodynamic nature. It is indeed well known~\cite{Umezawa:1993yq} that within the scheme one can consistently define  thermodynamic quantities (operators) such as the entropy and the free energy. Let us consider here the entropy (for remarks on the free energy see Appendix C).

Thinking of the entropy as an ``index'' or a measure of the degree of ordering present in the state of the system (lower entropy corresponding to higher degree of order), one can show that  the state $|0 (\theta) \rangle_{\cal N}$, which as we have seen is characterized by  a given value of the order parameter,
can be constructed by the use of the entropy operator $S$~\cite{Celeghini1992,BJV,Umezawa:1993yq}. The value of $S$ in $|0 (\theta) \rangle_{\cal N}$ is given by its expectation value in the familiar form~\cite{Celeghini1992,Umezawa:1993yq}:
\be\lab{entr3} {}_{\cal N}\langle 0(\theta) |S|0(\theta) \rangle_{\cal N} =
\sum_{n=0}^{+\infty} W_n \; \log W_n ~, \ee
where $W_n \equiv W_n (\theta) $ is given in the Appendix C.

Remarkably and consistently with the breakdown of time-reversal symmetry in dissipative systems (the appearance of the {\it arrow of time}), as commented in subsection II.A,
time evolution can be shown to be controlled by the entropy variations~\cite{Celeghini1992,DeFilippo:1977bk}. Indeed one finds that variations of the entropy control the variations in the $A-{\tilde A}$ content of  $|0 (\theta) \rangle_{\cal N}$, thus controlling the time evolution (the trajectories) in the manifold of concepts (the space of the {\it infinitely many} LF, see Heim in footnote 17). Entropy is thus related with the semantic level of the LF, {\it meanings}, which are dynamically arising as collective modes out of the syntactic (atomistic) level of the basic lexical elements.

In conclusion, we have uncovered the isomorphism between the  physics of many-body systems and the linguistic strategy of the Minimalist Program. However, although we have exploited the algebraic properties of the many-body formalism, in our scheme the linguistic structures are ``classical'' ones. It is known, on the other hand, that the many-body formalism  is well suited to describe not only the world of elementary particle physics and condensed matter physics, but also macroscopically behaving systems characterized by ordered patterns~\cite{BJV,Umezawa:1993yq}.
It is an interesting question whether the crucial mechanism of the foliation of the space of the states has to do with the basic dynamics underlying the linguistic phenomena observed at a macroscopic level. It might well be possible that the basic dynamics underlying the richness of the biochemical phenomenology of the brain behavior~\cite{Vitiello:1995,Vitiello:2001,FreemanVitiello:2006,FreemanVitiello:2008,CapolupoPLR}, also provides the basic mechanisms of linguistics.

\appendix

\section*{Appendix A}

\section*{A little bit of history}

(1) {\it Phrase Structure Grammars}

Until the mid-Fifties, a ``grammar'' was the result of applying some procedures (typically, segmentation and categorization) to a corpus of data\footnote{The most sophisticated development procedures were those of Zellig S. Harris
(Chomsky's teacher, at the time)~\cite{Harris}.}. 
Chomsky's first attempt at a new perspective was his master
thesis~\cite{Chomsky1951,Chomsky1979}, 
but did not reach any audience until {\it Logical Structure of Linguistic Theory} (1955), his ``Three models'' paper in 1956~\cite{Chomsky1956} 
and {\it Syntactic Structures} in 1957~\cite{Chomsky1957,Lasnik}.
Before the advent of his ``transformational grammar'', the dominant models were Phrase Structure Grammars (PSG) a usually long list of general rules specifying how each kind of Phrase (each phrasal constituent) was to be expanded, all the way to the manifest expression of a complete sentence. The arrows indicate ``rewriting rules'', that is: what appears on the left has to be rewritten as what appears on the right. For instance (a very simple one) for a whole sentence (S), we typically have as constituents a Noun Phrase (NP), a Verb Phrase (VP) and (optionally) a Prepositional Phrase (PP)

\vspace{0,3cm}

S $\rar$ NP VP NP (possibly also PP)

NP = Noun Phrase

VP = Verb Phrase

PP = Prepositional Phrase

\vspace{0,3cm}

More concretely

NP1 $\rar$  the cat

AUX $\rar$  is

VP $\rar$  (is) chasing

NP2 $\rar$  a mouse

PP $\rar$   in the garden

(S) = The cat is chasing a mouse in the garden.

\vspace{0,4cm}

(2) {\it Enter Transformational Grammar}

Chomsky did persuasively show that such grammars were radically insufficient to account for real languages. That there was a {\it kernel} sentence that can be {\it transformed} into variants that are intimately related, by permuting the order of the words in the phrases (obtaining passives, interrogatives, negations etc.). In his work {\it The Logical Structure of Linguistic Theory}  LSLT~\cite{Chomsky1955},
which really was also the first complete phrase structure grammar, kernel sentences were derived by obligatory transformations from a more abstract phrase structure.  E.g., a transformation was required to convert [[be-ing] chase] to [be chasing] (and another similar one to yield is), what Haj Ross later called ``Affix-hopping''. Suitable mathematical analyzes, in terms of the Theory of Automata and the theory of recursive functions, did show that Transformational Grammars (TG) constitute a more powerful kind of automata than Phrase Structure Grammars~\cite{Chomsky1959}. 
Some authors, on the basis of mathematical analysis, argued that TGs were ``too powerful'', because they over-generate, i.e. generate all the well-formed expressions but also expressions that are not syntactically well-formed~\cite{Peters1971,Peters1973}.
Chomsky's counter-critiques were twofold: over-generation follows trivially from the fact that TGs as such had no conditions on erasure\footnote{This was well corrected in the following years, first by imposing constraints on transformations, then by characterizing ``islands'' for extraction~\cite{Ross1984,Ross1967,Ross1986}
and then by the notion that certain syntactic categories in certain configurations are ``barriers'' to government and to movement~\cite{Chomsky1986}.  
In the different conceptual framework of today's Minimalism these restrictions follow from more abstract and simpler principles.}
Moreover, the empirical import of the issue was doubtful, because these analyses are concerned with ``weak generative power'' (producing acceptable surface expressions, no matter how), while the real concern is with ``strong generative power'' (explaining how the set of structural descriptions, such as Merge, derivations, trees, proof trees, is assigned by a formal system to the strings that it specifies). The only empirically significant concept in Generative Grammar is strong generative capacity.
Canonical examples of transformations are like the following:

\vspace{0,3cm}

{\it  The cat is not chasing a mouse in the garden}.   NEGATION

{\it  A mouse is being chased by the cat in the garden}.  PASSIVE

{\it  What/who is the cat chasing in the garden?} ACTIVE INTERROGATIVE

{\it  What/who is being chased by the cat in the garden?}  PASSIVE INTERROGATIVE

{\it  What/who is not being chased by the cat in the garden?}  NEGATIVE INTERROGATIVE
\vspace{0,3cm}

Instead of having many more ad hoc phrase structure rules, transformations took care of such cases\footnote{Leaving inessential details aside and stressing that negation has a straightforward representation in formal logic, all these sentences have the same Logical Form (provided we  insert or do not insert the logical symbol for negation, as required).}.

The kernel, then was generalized into a deep structure distinct from the surface structure. The link with meaning was supposed to be the deep structure, not the surface structure. All speakers understand that the basic syntactic and semantic relations are preserved in sentences such as

\vspace{0,3cm}
{\it  He met his friend}.

{\it  He succeeded in meeting his friend}.

{\it  He failed to meet his friend}.

{\it  He avoided meeting his friend}.
\vspace{0,3cm}

Separate ad hoc phrase structure rules would have to be introduced, while basic transformational rules cover all such construction and explain why the basic meaning is invariant at the level of the deep structure\footnote{We will not detail here how modifier verbs like ``succeed'' and ``fail'' are easily expressed in Logical Form.}. Also, there are physically absent, but mentally present, syntactic elements that PSG\footnote{Phrase Structure Grammars} rules could not possibly deal with. Between square parentheses what is not pronounced (or written)

\vspace{0,3cm}
{\it  He saw the man standing at the bar}.

{\it  He saw the man [the man was] standing at the bar}.

{\it  They saw the prisoner escape from the prison}.

{\it  They saw the prisoner [when they saw the prisoner he was] escap(ing) from the prison}.
\vspace{0,3cm}

Importantly, there are verbs that block such constructions\footnote{Lexical semantics explains why these differences exist. Simplifying drastically, verbs like ``suspect'' (called Psych Verbs) cannot take a gerundive form as their direct object, while verbs of perception like ``see'' can. }. Ill-formed sentences are, by a long tradition, preceded by an asterisk.
\vspace{0,3cm}

{\it  They suspected that the prisoner would try to escape}.

{\it  *They suspected the prisoner trying to escape}.

Contrast this with the perfectly OK sentence

{\it They saw the prisoner trying to escape}.
\vspace{0,3cm}

So, silent elements were introduced for the first time in linguistics. They are silent (``empty'' in the technical terminology) but mentally present and very important in the syntactic structure and in the interpretation.

\vspace{0,4cm}

3) {\it  The Theory of Government and Binding}

In subsequent years, syntactic movement became a key notion, more general than transformations. Elements in a sentence can be moved to a different position, leaving a trace (an ``empty category''). The interpretive system understands the meaning of the sentence by (mentally, silently) paying attention to the position of the trace and to what ``governs'' the trace.
\vspace{0,3cm}

{\it  Which books did you read?}
\vspace{0,3cm}

Books are the object of read and are so interpreted, as being to the right of read, where the trace is. Making the trace explicit by the letter t, we have
\vspace{0,3cm}

{\it  Which books did you read t?}

\vspace{0,3cm}

 Making syntactic movement even more explicit, we can write:
\bea
Which&&books~ did~ you~ read ~t? \non\\
       && \uparrow_{\leftarrow --\leftarrow--\leftarrow--\leftarrow--}  \downarrow  \non
 \eea

\vspace{0,3cm}

Several kinds of un-pronounced (and un-written) syntactic elements were introduced, called ``empty categories'' (empty because un-pronounced). But they had to be governed in a very similar way to the explicit elements of the sentence (this was called ECP, the Empty Category Principle).

Government was exactly defined, but we will not go into it here, because this notion has been subsumed by a deeper, more abstract notion in the current Minimalist version of the theory.

It's worth noticing that there are ``barriers'' to syntactic movement, that is: some instances of syntactic movement are blocked. There are ``islands'' to extraction and movement.
\vspace{0,3cm}

{\it  That the police would arrest several rioters was a certainty.}

{\it  *Who was that the police would arrest a certainty?}
\vspace{0,3cm}

In the Theory of Government and Binding, these barriers for movement and the nature of ``islands'' were explained in terms of strict locality. Summarizing and simplifying drastically a number of acute analysis and several refinements, we can indicate here the so-called Minimal Link Condition~\cite{Chomsky1995}. 
Derived from general principles of derivational economy, it stated that movement can only occur to the closest potential landing site.

A similar situation occurs, for instance, with agreement. In Italian there are masculine and feminine features for inanimates: {\it la bella sedia}, [FEM]  {\it il grande coltello} [MASC], also plural/singular for determiners and adjectives ({\it le belle sedie, i grandi coltelli}).  Once the noun gives the interpretable information singular (only one) or several (plural), these agreement features on the determiners and the adjectives are redundant (in fact English does not have them). For reasons of economy, Narrow Syntax must delete all un-interpretable features {\it before} the sentence reaches the Conceptual-Intentional interface.
Basically, the process was the following: uninterpretable features must be {\it checked} (sic) to be deletable. Syntactic Objects with such features must ``check'' them with Syntactic Objects that have interpretable features. The search is {\it strictly local}, as local as possible. Best: inside the same constituent, at worst: inside the same Phase. The process is called probe-goal. When the probing Syntactic Object encounters the goal, the uninterpretable features are deleted and the
result is sent to the Conceptual-Intentional interface. If the local search fails, we have an ungrammatical sentence (the derivation is said to ``crash'').
Uninterpretable features that have been successfully checked are deleted at the Conceptual-Intentional interface, but not necessarily (in fact usually not) also at the Sensory Motor interface. In fact we do see and hear them at the level of pronunciation and writing. In {\it Le molte belle figurine} (the many beautiful little cartoons) feminine and plural features are visible and audible in the determiner, the quantifier and the noun. This is called Agreement, the object of many, many linguistics studies~\cite{Baker2013a,Baker2013b,Miyagawa2010}. 
Agree is what makes determiners in languages such as Italian ({\it il, lo , la, molto, molti, poco, pochi} etc.) and adjectives ({\it bello, bella, belli}) agree with the noun. Also what makes auxiliaries ({\it have, be}) and past participles agree with the tense of the verb ({\it era stato visto, furono visti, saranno stati comprati} etc.). Agree must be local, very local, in a structural sense, not necessarily ``local'' in terms of proximity of words in the surface structure. I {\it libri di cui mi parlasti erano stati tutti comprati in libreria} (the books you had told me about have all been bought in a bookshop). {\it Libri} and {\it comprati} are not ``near'' on the surface, but they are near structurally, because the Prepositional Phrase {\it di cui mi parlasti}, which contains a Verb Phrase, veers off (so to speak) into a sideway structure (it's an adjunct). A similar constraint applies to the assignment of Case (nominative, accusative, genitive etc.)~\cite{Baker2013a,Baker2013b}.  
Case assigners (verbs and prepositions) must be structurally local to the nouns, adjectives etc. to which they assign case. Again, locality is structural, not necessarily on the surface. (Let's think of Latin, Russian etc. where Case is overtly expressed).

\section*{Appendix B}

\section*{Some useful formulas on Fibonacci matrix}

The matrix
\bea
F \equiv \frac{1}{2}\,I  + \sigma_3 + 2 \, \sigma_1 =
 \frac{1}{2}\,\left(
  \begin{array}{cc}
    1 & 0 \\
    0 & 1 \\
  \end{array}
\right) + \frac{1}{2}\left(
  \begin{array}{cc}
    1 & 0 \\
    0 & -1 \\
  \end{array}
\right) +  \left(
  \begin{array}{cc}
    0 & 1 \\
    1 & 0 \\
  \end{array}
\right) = \left(
  \begin{array}{cc}
    1 & 1 \\
    1 & 0 \\
  \end{array}
\right) \non
\eea
is called the Fibonacci matrix. For the $n$-powers $F^n$ of the $F$ matrix, with $n \neq 0$, we have
\bea
F^{n} &=& \left(
  \begin{array}{cc}
    F_{n+1} & F_{n} \\
    F_{n} & F_{n - 1}\\
  \end{array}
\right) = F_{n - 1} \,I + F_{n} \, F,  \qquad {\rm n} \neq 0  \non
\eea
where the elements $a_{ij}$, $i, j = 1, \,2$,  in the matrix $F^n$ reproduce, in the order shown in the equation above, the numbers $F_{n+1},  \,F_{n}, \, F_{n}, \, F_{n - 1}$, with $F_{0} \equiv 0$, in the Fibonacci progression
$0,\, 1, \,1, \,2, \,3, \,5, \,8, \,13,....$. Moreover, the coefficients of the matrices $I$ and $F$ in the last member on the r.h.s of the above relation also reproduce the  numbers  $F_{n - 1}$ and $F_{n}$, respectively, in the Fibonacci progression. We can indeed verify that
\bea
F^{1} &=& \left(
  \begin{array}{cc}
    1 & 1 \\
    1 & 0 \\
  \end{array}
\right) =  F, \non \\
F^{2} &=& \left(
  \begin{array}{cc}
    2 & 1 \\
    1 & 1 \\
  \end{array}
\right) = I + F, \non \\
F^{3} &=& \left(
  \begin{array}{cc}
    3 & 2 \\
    2 & 1 \\
  \end{array}
\right) = I + 2\, F,\non \\
F^{4} &=& \left(
  \begin{array}{cc}
    5 & 3 \\
    3 & 2 \\
  \end{array}
\right) = 2\,I + 3\, F,\non \\
F^{5} &=& \left(
  \begin{array}{cc}
    8 & 5 \\
    5 & 3 \\
  \end{array}
\right) = 3\,I + 5\, F,\non \\
F^{6} &=& \left(
  \begin{array}{cc}
    13 & 8 \\
    8 & 5 \\
  \end{array}
\right) = 5\,I + 8\, F,\non \\
......&etc.&.....~, \non
\eea

\section*{Appendix C}

\section*{On the entropy and the free energy}

The state $|0 (\theta) \rangle_{\cal N}$ can be constructed by the use of the entropy operator $S_A$~\cite{Celeghini1992,BJV,Umezawa:1993yq}. In fact we can write:
\bea \non \lab{entrVac}
|0 (\theta) \rangle  = \exp{\left ( -
{1\over{2}} S_{A} \right )} |\,{\cal I}\rangle  \, = \exp{\left (
- {1\over{2}} S_{\ti A} \right )} |\,{\cal I}\rangle  ~,
\eea
where
\bea \non
 S_{A} \equiv  - \sum_{\bf k} \lf( A_{\bf k}^{\dagger} A_{\bf
k} \log \sinh^{2} {\theta}_k - A_{\bf k} A_{\bf k}^{\dagger} \log
\cosh^{2} {\theta}_k \ri)~, 
\eea
is the entropy operator and $ |\,{\cal I}\rangle \, \equiv \exp {\left( \sum_{\bf k}
A_{\bf k}^{\dagger} {\ti A}_{\bf k}^{\dagger} \right)} |0\rangle $.
 $S_{{\ti A}}$ has an expression similar to $S_{A}$, and is obtained by replacing
$A$ with ${\ti A}$.  $S$ will denote either $S_{A}$ or $S_{{\ti A}}$.
The state $|0(\theta)\rangle$, where now for simplicity we omit the subscript $\cal N$,  can be also written as~\cite{Celeghini1992,Umezawa:1993yq}:
\bea 
\non  |0 (\theta) \rangle = \sum_{n=0}^{+\infty} \sqrt{W_n} \left( |n \rangle
  \otimes |{n} \rangle  \right) ~,
\eea
with
\be \non
  W_n = \prod_{\bf k}
  \frac{\sinh^{2n_{\bf k}}\theta_k}{\cosh^{2(n_{\bf k}+1)}\theta_k}\,,
\ee
and $n$ denoting the set $\{ {n}_{\bf k} \}$,  $ 0 < W_n
< 1$, $~\sum_{n=0}^{+\infty} W_n = 1$. The entropy acquires then the form given in Eq.~(\ref{entr3}).

The free energy ${\cal F}_{A}$ can be
defined~\cite{Umezawa:1993yq,Celeghini1992}
for $|0(\theta) \rangle$:
\be \non 
{\cal F}_{A} \equiv \langle 0(\theta)| \Bigl (H_{A}
- {1\over{\beta}} S_{A} \Bigr ) |0(\theta)\rangle~, \ee
where  $H_{A} = \sum_{\bf k} \hbar \omega_{k} A_{\bf
k}^{\dagger} A_{\bf k}$.  By defining  heat as usual ${dQ={1\over{\beta}} dS}$,  from the stationarity condition $ d {\cal F}_{A} = 0$ we have
\be \non 
d E_{A} = \sum_{\bf k} \hbar \,\omega_{k} \,
\dot{{\cal N}}_{A_{\bf k}}(t)d t = {1\over{\beta}} d {\cal  S} = d Q ~, \ee
which relates the time derivative ${\dot{\cal N}}_{a_{\bf k}}$  and entropy variations (heat dissipation $dQ$).

\section*{Acknowledgements}
We are grateful to Noam Chomsky for many constructive suggestions concerning our summaries of the relevant parts of linguistic theory. We are also grateful to Walter J. Freeman for useful and inspiring discussions. Illuminating discussions with Gianfranco Basti are also acknowledged.
We express our immense gratitude to the dear memory of our friend and colleague, the late Emilio Del Giudice, who initiated with us this research and gave many invaluable insights, too many and too deep for us to be able to
single them out and report them in a specific way.
Miur and INFN are acknowledged for partial financial support.

\end{document}